\documentclass[letterpaper, 10 pt, conference]{ieeeconf}  

\IEEEoverridecommandlockouts                              

\usepackage{graphicx}
\usepackage[linesnumbered,ruled,vlined]{algorithm2e}
\usepackage[noend]{algpseudocode}

\usepackage{amsmath}
\usepackage{amssymb}
\usepackage{multirow}
\usepackage{booktabs}
\usepackage{float}
\usepackage{soul}
\usepackage{xcolor}
\usepackage{pifont}

\newcommand{\xmark}{{\large \ding{55}}}
\sethlcolor{yellow}

\usepackage[noadjust]{cite}

\title{\LARGE \bf
MultiTalk: Introspective and Extrospective Dialogue for Human-Environment-LLM Alignment
}

\author{
    Venkata Naren Devarakonda$^{*1}$, Ali Umut Kaypak$^{*1}$, Shuaihang Yuan$^{2}$, \\ Prashanth Krishnamurthy$^{1}$, Yi Fang$^{2}$, Farshad Khorrami$^{1}$
    \thanks{*Equal Contribution}
    \thanks{$^{1}$Department of Electrical and Computer Engineering, NYU Tandon School of Engineering, Brooklyn, NY, 11201}
    \thanks{$^{2}$Department of Electrical and Computer Engineering, New York University Abu Dhabi, Abu Dhabi 129188, UAE}
}

\begin{document}

\maketitle
\thispagestyle{empty}
\pagestyle{empty}

\begin{abstract}

LLMs have shown promising results in task planning due to their strong natural language understanding and reasoning capabilities. However, issues such as hallucinations, ambiguities in human instructions, environmental constraints, and limitations in the executing agent’s capabilities often lead to flawed or incomplete plans. This paper proposes MultiTalk, an LLM-based task planning methodology that addresses these issues through a framework of introspective and extrospective dialogue loops. This approach helps ground generated plans in the context of the environment and the agent's capabilities, while also resolving uncertainties and ambiguities in the given task. These loops are enabled by specialized systems designed to extract and predict task-specific states, and flag mismatches or misalignments among the human user, the LLM agent, and the environment. Effective feedback pathways between these systems and the LLM planner foster meaningful dialogue. The efficacy of this methodology is demonstrated through its application to robotic manipulation tasks. Experiments and ablations highlight the robustness and reliability of our method, and comparisons with baselines further illustrate the superiority of MultiTalk in task planning for embodied agents.
\end{abstract}

\section{Introduction}

\label{sec:intro}
With the advancement of Large Language Models (LLMs) (\cite{dubey2024llama3,openai2023gpt4,anil2023palm}), there is a growing interest in generating task plans for robots by leveraging the reasoning ability of LLMs (\cite{huang2022language, pmlr-v205-huang23c, ahn2022can, 10161317, 10160591, sun2024adaplanner}). However, initial attempts lacked accuracy due to LLMs' hallucinations (\cite{ji2023survey,rawte2023survey,tonmoy2024comprehensive,liu2024survey}), which resulted in incorrect plans. To enhance planning capabilities, previous methods employed multiple LLMs or visual language models (VLMs) as critics to refine and update the initial plans (\cite{stechly2024self,renze2024self}). Nonetheless, naively increasing the number of LLMs and assigning different LLMs as experts for various tasks (\cite{long2023discuss,feng2024improving}) or using additional LLMs as validators for plan refinement \cite{NEURIPS2023_1b44b878} have proven to be less effective and computationally heavy (\cite{kambhampati2024llms,valmeekam2023planning}). To address ambiguities or incorrect plans, direct human intervention for feedback correction during task execution is considered (\cite{sharma2022correcting,pmlr-v205-huang23c}). This approach, while reducing the system's overall automation, significantly increases the reliability and safety of task execution, which is especially vital in complex scenarios where AI alone might not suffice.

Potential dangers that need to be addressed to enable LLMs to become a versatile solution for task planning for robots include LLMs' propensity to hallucinate, inaccurate responses, ambiguity and imprecision in the natural language used by users to communicate task instructions, and uncertainties in the environment and in the robot's capabilities and constraints. Hence, a critical step towards a truly effective solution would be to introduce robust feedback mechanisms that can identify and address these issues. Further, automating these feedback mechanisms would reduce human involvement while improving planning performance.

\begin{figure*}[t]
      \centering
      \includegraphics[width=2\columnwidth]{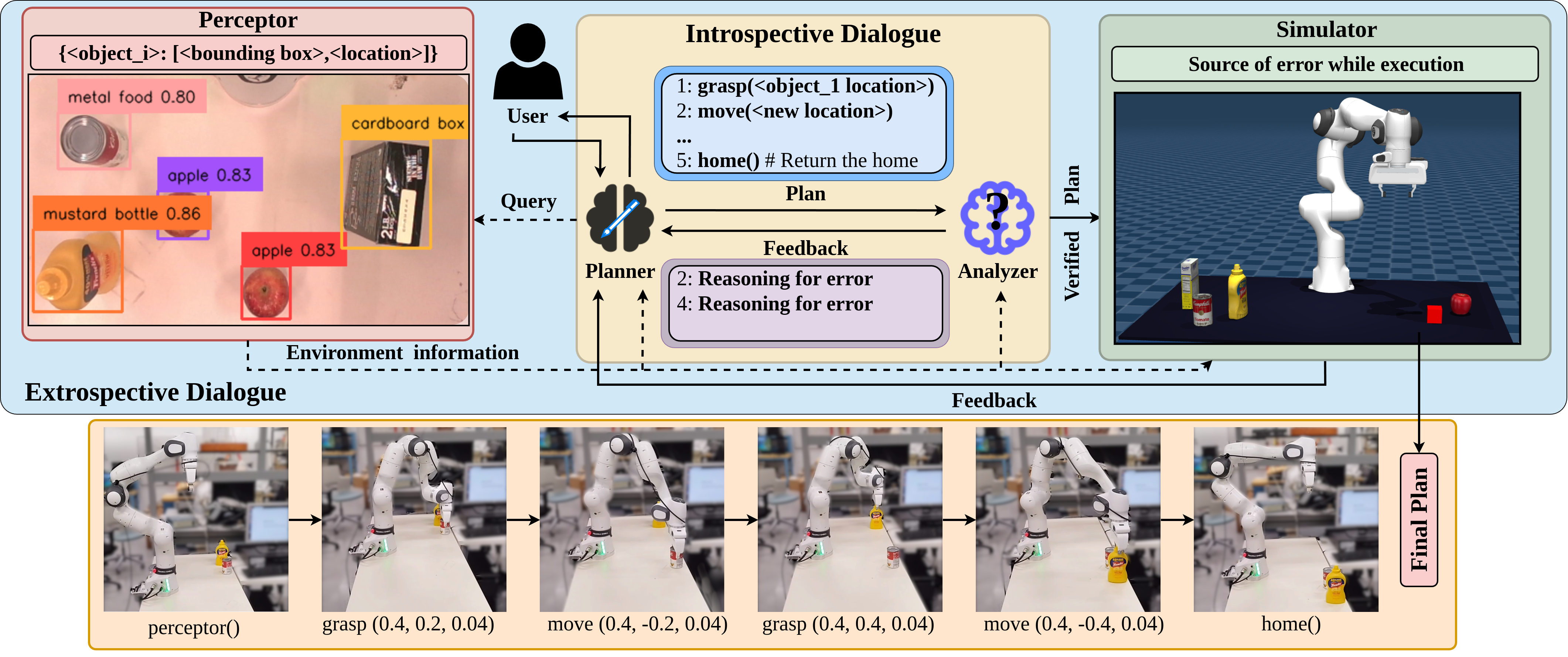}
      \caption{Diagram illustrating the interaction and feedback loops between the four main modules of MultiTalk: Perceptor, Planner, Analyzer, and Simulator. Dashed lines indicate the flow of environmental data and feedback, while solid lines represent the flow of plans and plan-related feedback. The Perceptor identifies objects in the scene and informs the other modules. The Planner generates the plan and interacts with the other modules to disambiguate the task and iteratively refine the plan. The Analyzer critiques the Planner's output, while the Simulator ensures the plan is grounded in the robot's capabilities and environmental constraints. Finally, the refined plan is executed by the robot.}
      \label{fig:diagram}
\end{figure*}

Based on this idea, we propose MultiTalk, a method that enables introspective and extrospective dialogue to accurately break down tasks into executable plans  (see Figure \ref{fig:diagram}). Our contributions in this paper include: 
\begin{enumerate}
    \item An LLM-based task planner with feedback systems that track user intentions, environmental constraints, and robot capabilities, using a novel introspective and extrospective dialogue framework
    \item Specialized modules that detect and resolve hallucinations and logical errors, clarify ambiguities in input tasks, and observe plan-dependent temporal states for executability on the robot
    \item Demonstrating the effectiveness of MultiTalk through ablation studies, baseline comparisons, and real-world implementation on a robotic manipulator
\end{enumerate}

\section{Related Work} \label{sec:rw}
Recent advancements in LLMs and VLMs have significantly enhanced planning and decision-making processes in robotics, including some methods that have finetuned foundation models to create Vision-Language-Action models that directly predict robot actions (\cite{openvla,li2023vision,huang2023embodied,zhen20243dvla,zhang2024sam}). However, many studies rely heavily on the in-context capabilities of LLMs (\cite{brownfewshot,wei2022cot,guan2023leveraging,yao2023react}). For instance, Huang et al. \cite{huang2022language} utilize LLMs to decompose robotic task objectives into a sequence of high-level instructional actions that an agent can execute. Beyond generating natural language descriptions, Code-As-Policies \cite{10160591} and ProgPrompt \cite{10161317} propose directly generating actions in executable code, therefore improving execution precision. Additionally, recent approaches suggest leveraging visual scene information to generate plans, employing either VLMs (\cite{navgpt,lmnav,voxposer,hu2023look}) or object detectors (\cite{text2motion,shirai2024vision}). Despite the promising results achieved for various robotic planning tasks, these methods often lack robust correction mechanisms to address errors during execution, which can limit their practical applicability.

We present a framework that improves existing task planning solutions by addressing their limitations. For example, SayCan {\cite{ahn2022can}} grounds LLM-based plans in the environment by choosing the most affordable action based on camera input, but it lacks feedback systems to correct logical errors and practical feasibility issues. In contrast, we implement introspective dialogue that ensures correction of grounding-based, hallucination, and logical errors. Reflexion {\cite{NEURIPS2023_1b44b878}} attempts to address these issues by using an LLM-based Reinforcement Learning methodology: separate instances of LLMs are used to generate plans and observe the output and associated reward. This reward and additional reflective text generated by LLMs are used to build short-term and long-term contextual memory for plan correction. ISR-LLM \cite{10610065} employs a second LLM validator agent, similar to the Analyzer in MultiTalk. However, these methods do not provide a solution for grounding the plan into feasible robot motions, a challenge we address by incorporating a simulator.

One of the works most closely related to ours is Inner Monologue \cite{pmlr-v205-huang23c}. It uses human feedback to resolve ambiguity by leveraging various feedback sources for environmental grounding and binary detectors to assess task executability on the robot. However, relying solely on binary detectors limits iterative improvement by preventing meaningful, directed feedback, often resulting in repeated failures. MultiTalk addresses this by using specialized feedback systems that observe the robot in a simulator, providing precise failure reasoning and potential corrections. Additionally, Inner Monologue's success detector requires executing the plan, which risks damaging the robot—a risk we avoid by incorporating a simulator into the framework.

\begin{algorithm}
\small
    \SetAlgoLined
    \textbf{Parameters}: max\_iterations \\
    \KwIn{instruction, detected\_objects}
    \textbf{Modules}: planner, analyzer
    \tcp{Initialization}
    $i \gets 0$\;
    $\text{feedback\_source} \gets \text{None}$\;
    $\text{feedback} \gets \text{None}$\;
    \While{$i$ < \textnormal{max\_iterations}}{
        planner\_output $\gets$ planner.call(instruction, feedback, feedback\_source)\; 
        \uIf{ \textnormal{planner\_output.type = `perception feedback' }}{
        
            feedback $\gets$ get\_updated\_objects() \Comment{re-scan environment}
            feedback\_source $\gets$ `perception'\;
        }
        \uElseIf{ \textnormal{planner\_output.type = `question human' }}{
        
            feedback $\gets$ get\_answers\_from\_human(planner\_output.questions) \Comment{answers}
            feedback\_source $\gets$ `human'\;
        }
        \ElseIf{ \textnormal{planner\_output.type = `instructions' }}{
            feedback $\gets$ check\_workspace\_bounds(planner\_output.plan)\;
            feedback\_source $\gets$ `analyzer'\;
            \If{ \textnormal{`out of bounds' in feedback }}{
                continue\;
            }
            feedback $\gets$ analyzer.call(instruction, planner\_output.plan)\Comment{analyzer feedback}
            \tcp{Once accepted by analyzer, the simulator checks}
            \If{ \textnormal{feedback = `feasible' }}{
                feedback $\gets$ run\_simulation(planner\_output.plan, detected\_objects) \;
                \uIf{\textnormal{ feedback == `success' }}{
                    break \Comment{Plan is feasible!}
                }
                \Else{
                feedback\_source $\gets$ `simulator' \Comment{send simulator feedback}
                }
                
            }
        }
        $i \gets i+1$\;
    }
    \Return{\textnormal{planner\_output.plan}}
    \caption{MultiTalk}
    \label{alg:1}
\end{algorithm}

\section{Methodology} \label{sec:method}

\subsection{Problem Formulation}

In this paper, we seek to develop a methodology to translate a high-level natural language instruction/task into a plan containing a sequence of task-specific primitives, such that the generated plan aligns with the intentions of the user, the constraints of the operational environment, and the capabilities of the robotic agent performing the task. Here, \textit{alignment} in the context of the user indicates whether the details of the plan are as per the user's request and expectations. Human value alignment is not considered or studied in this work. With regard to the environment and the executing agent, alignment is equivalent to feasibility. This alignment is achieved through feedback from multiple sources, each monitoring different aspects of the alignment. While the approach requires seamless interaction with the user when necessary, to clarify ambiguities in the task, it does not expect the user or any other source to provide completely infallible feedback. We present a solution that harnesses the LLM's generative capabilities and ability to be conditioned with natural language for task planning while mitigating its downsides such as hallucination and logical flaws.

\subsection{Introspective and Extrospective Dialogue}

To achieve plan alignment, the proposed approach uses multiple feedback loops arranged into introspective and extrospective dialogue channels, as seen in Figure \ref{fig:diagram}. Following the flow depicted by Algorithm \ref{alg:1}, a feasible plan is output after multiple iterations of dialogue. The extrospective dialogue comprises mechanisms that observe the external environment and the robot's states throughout the execution of the plan, identify possible errors, flag them, and convey them through natural language feedback to the planning agent. This is enabled by three main systems: the Perceptor, Simulator, and User. The Perceptor updates the other modules with information on the objects present in the workspace, their locations, and bounding boxes. The Simulator observes additional object and robot states to detect features like collisions, singularities, and controller faults while executing the plan refined through introspective dialogue.

The introspective dialogue involves the interaction between two LLM agents, the Planner and the Analyzer, to autonomously evaluate and critique the generated plans considering the information and feedback from the other modules. The Planner is responsible for generating a feasible plan for a given task, aligning with feedback from multiple sources. The Analyzer, an LLM-based module similar to the Planner, critiques the plan and checks if it is consistent with the presented feedback. It also helps identify and correct hallucination, logical, and syntactic errors in the LLM-generated plans. Another key aspect of the introspective dialogue is to evaluate the input task and detect instances where the user instructions are unclear, imprecise, or incompatible with the environment state as seen by the Perceptor. It then triggers an extrospective conversation with the user or the Perceptor to attempt resolution through appropriate clarifying questions or by requesting additional sensor data.

\begin{figure}
    \centering
    \includegraphics[width=1\linewidth]{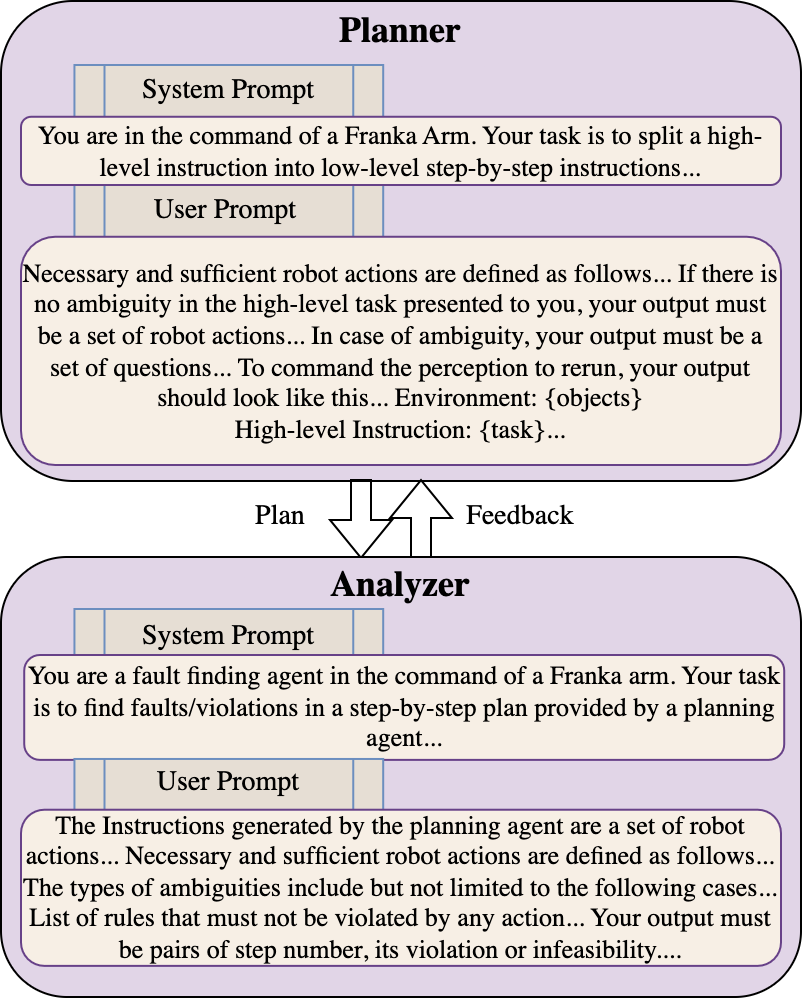}
    \caption{Overview of the system and user prompts for the LLM agents in the framework. The LLM agents are assigned distinct roles via tailored system and user prompts. The Planner generates a robotic plan including actions and their corresponding arguments, while the Analyzer identifies potential errors in the plan.}
    \label{fig:prompts}
\end{figure}

\subsection{The Perceptor}

This module utilizes an Intel RealSense D455 camera to capture an RGB-D image of the scene and perform bounding box detection, labeling, and closed-set segmentation of all the objects using the Grounded SAM \cite{ren2024grounded} framework. Combined with the depth information, this results in a list of objects along with their 3D bounding box dimensions and centers, which it feeds to all the other systems.

We include a Perception module in our proposed framework instead of utilizing an oracle to make our system more suitable for real robot scenarios. Most of the time, errors such as misclassification and failure to detect objects can be corrected by altering the camera’s viewpoint when the Planner is in the loop. When an error occurs—such as when the user requests an item but it is not detected—the Planner instructs the robot arm to adjust its camera position, capturing the environment from a different angle. This additional information also helps the Planner decide whether the Perceptor made an error or if the input task is infeasible, enabling it to generate feasible plans accordingly. 

\subsection{The Planner}

The Planner is an instance of the GPT-4o LLM, equipped with carefully designed prompts for handling feedback from other components within the framework. After receiving instructions from the user and scene information from the Perceptor, its primary duty is to develop a plan based on the action primitives defined in the prompts. Additionally, when instructions are ambiguous, the Planner must seek clarification by asking the user specific questions, such as the target location of requested objects. A crucial aspect of our Planner's design is its ability to strike a balance: minimizing the number of questions posed to the user while ensuring the instructions are unambiguous.

\subsection{The Analyzer}

This module functions as a critique agent to the Planner. Like the Planner, it is a separate GPT-4o model instance with the same environmental information, but with different prompts (see Figure \ref{fig:prompts} for an overview), instructions, and additional heuristics to detect physical environment violations that may occur in the plans. Additionally, it is commanded to check for logical, syntactic, and hallucination errors in the plan. The prompts to this model are focused on how to interpret input plans, assess them for faults, and handle disagreements between itself and the Planner. To enable logical dialogue between the agents, the Analyzer is prompted to provide sufficient reasoning in its feedback. The Planner and the Analyzer engage in this dialogue in a loop until they agree upon a feasible plan.

\subsection{The Simulator}

This MuJoCo-based \cite{todorov2012mujoco} dynamic simulation  module ensures that the proposed plan adheres to physical constraints. The Planner's actions may cause collisions with objects or violate robotic constraints like joint limits and singularities, as LLMs are incapable of comprehending complex physical constraints posed by the robot dynamics and its environment. The Simulator creates a virtual environment using information from the Perceptor and runs the plan approved through the introspective dialogue, checking for such errors. Further, by employing the same low-level controller as the real robot, the Simulator ensures that the trajectory can be accurately executed by the real robot. 

A threshold on the condition number of the manipulator Jacobian during the simulation helps prevent singularities. Contact between objects can be observed in MuJoCo during simulation to check for collisions and provide optimal feedback to the Planner. Finally, separate feedback is provided for cases when the controller fails to move the object accurately to the desired position.

The aim of using a simulator is not to replicate the real scene exactly but to create an environment where errors such as collisions, joint limit violations, or motion through singularity points can be detected. This approach is effective in setups involving a closed set of objects for which simulator models are available. Additionally, the methodology remains applicable even when a precise 3D object model or its geometric approximation is unavailable. In such cases, the simulator can still be used to identify controller errors and manipulator singularities in the trajectories.

\begin{figure}
      \centering
\includegraphics[width=0.9\columnwidth]{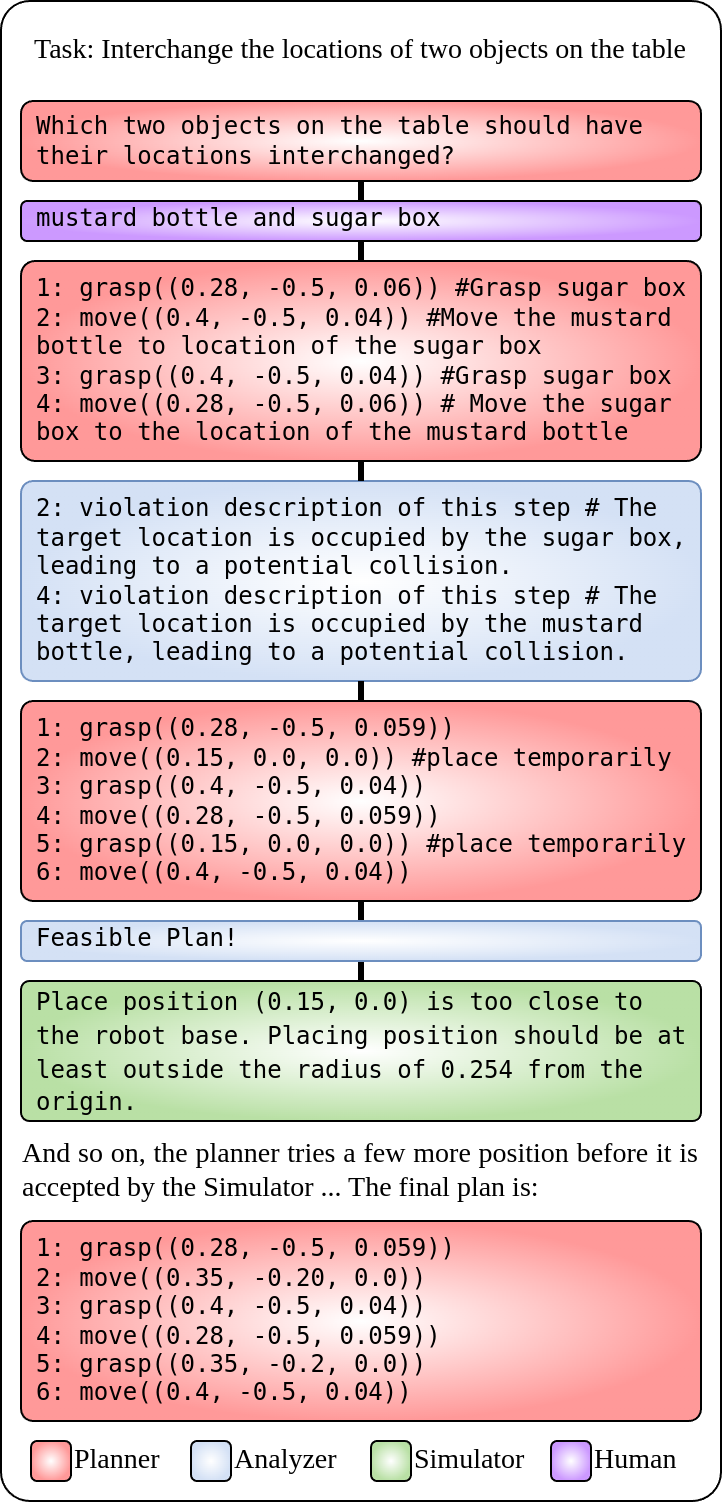}
      \caption{An example of the complete pipeline in action to generate a feasible plan. The Analyzer assists the Planner in realizing the correct logic and the Simulator helps in finding an approachable and unoccupied temporary location on the table. The importance of each component in resolving the logic and aligning the plan with the environmental and executional constraints are clearly seen here.}
      \label{fig: test case}
\end{figure}

\begin{table*}[t]
\caption{Comparison of Success Rate (\textit{SR}) between the overall method and its ablated versions. Each output plan's success is evaluated by manually checking for correctness and executability on FR3.}
\centering
\begin{tabular}{@{}lcccc@{}}
\toprule
\multicolumn{1}{c}{\multirow{2}{*}{\textbf{Natural Language Description of Task}}}                                                                                      & \multicolumn{4}{c}{\textbf{Success Rate}}                                                                                                                                                                                                 \\ \cmidrule(l){2-5} 
\multicolumn{1}{c}{}                                                                                                                                                    & \textbf{\begin{tabular}[c]{@{}c@{}}Overall \\ Framework\end{tabular}} & \textbf{\begin{tabular}[c]{@{}c@{}}Planner +\\ Analyzer\end{tabular}} & \textbf{\begin{tabular}[c]{@{}c@{}}Planner +\\ Simulator\end{tabular}} & \textbf{Planner} \\ \midrule
\textbf{T1}: Give me the \textless{}object\textgreater{}                                                                                                                                                     & \textbf{0.8}                                                          & \textbf{0.8}                                                                   &                        0.4                                                & 0.1              \\ \midrule
\textbf{T2}: Give me something to eat                                                                                                                                                & \textbf{0.9}                                                          & \textbf{0.9}                                                                   &                         0.4                                               & 0.4              \\ \midrule
\textbf{T3}: Move the objects to the other side of the table                                                                                                                         & \textbf{1.0}                                                          & \textbf{1.0 }                                                                  &                       \textbf{1.0 }                                                & \textbf{1.0}              \\ \midrule
\begin{tabular}[c]{@{}l@{}}\textbf{T4}: The objects of the same category must be in the same quadrant\\and the objects of different categories in different quadrants\end{tabular} & \textbf{0.7}                                                          & 0.4                                                                   &                         0.5                                               & 0.0              \\ \midrule
\textbf{T5}: Arrange the objects to form a square                                                                                                       &                         \textbf{0.8}                                  &                        \textbf{0.8}                                           &                               \textbf{0.8}                                         &        0.6   \\ \midrule
\textbf{T6}: Interchange the locations of two objects                                                                                                                   & \textbf{0.8}                                                          & 0.3                                                                   &                          0.5                                              & 0.0              \\ \midrule
\textbf{T7}: Interchange the locations of two object pairs                                                                                                                  & \textbf{0.8}                                                          & 0.1                                                                   &                         \textbf{0.8}                                               & 0.0              \\ \midrule
\textbf{T8}: Arrange the objects on the table such that they form a straight line                                                          & \textbf{0.4}                                                          & 0.3                                                                   &                         \textbf{0.4}                                               & 0.3              \\ \bottomrule
\end{tabular}
\label{tab:srtable}
\end{table*}

\section{Experiments and Results}
\label{sec:result}

\subsection{Experimental Setup}

We employ a 7 DoF robot arm, the Franka Emika FR3, which executes plans generated by MultiTalk using three action primitives: \textit{grasp}, \textit{move}, and \textit{home}. The Perceptor captures the environment with an Intel RealSense D455 RGB-D camera mounted on the manipulator. Eight test scenarios are specifically designed to evaluate the planning quality, ambiguity resolution, and grounding abilities of our proposed framework. We define five different environment configurations for each task, using various combinations of object types, number, and locations. For each configuration, we choose one or more objects from a list of five object categories, selecting with replacement: sugar box, soup can, wooden cube, mustard bottle, and apple. The meshes for these objects in the MuJoCo-based simulator are sourced from the YCB dataset \cite{7251504}. Finally, to account for the variance in the LLM's output, we run each experiment twice, leading to a total of 80 experiments. To limit the runtime of the algorithm, we restrict each experiment to a maximum of 10 feedback loops. Further, to underscore the importance of each module within our framework, we conduct ablation studies by systematically removing each module and evaluating the success rate. Table \ref{tab:srtable} encapsulates the combined results from a total of 320 experiments.

The output of LLMs is probabilistic, highly variable, and sensitive to input prompts \cite{loya2023exploring}. A proper quantitative analysis shows how stable the model is with the designed prompts and how robust it is to varying human input. We specifically test the robustness of our method by testing it in various environmental configurations. The consistently high success rate demonstrates the method’s stability under different conditions. Note that the success of each experiment is determined by manually evaluating the logic of the generated plan and checking if the execution of the plan is error-free and aligns with the original intentions of the user.

\subsection{Ablation Studies}
\subsubsection{Quantitave Analysis} 
\label{sec: self-abl}
The ablation studies highlight the magnitude of improvement each module brings to the overall method. It can be observed in Table \ref{tab:srtable} that our method improves the performance of the Planner significantly for most tasks. Tasks 4, 6, and 7 depict cases that challenge the LLM to realize a complicated logic and apply it to multiple objects. The Planner almost always makes logical errors in such cases, as can be seen from the 0 success rate in all these tasks. Having the Analyzer in the loop helps the Planner identify the logical errors and correct them. 

Another interesting observation is the failure of the Planner when asked to interchange object locations. It completely fails to realize that it must first move the object to a temporary location, rather than directly to the location of the other object. In tasks 6 and 7, the results of Planner+Analyzer are misleading as they indicate very little improvement brought by the Analyzer. However, in this task, the Analyzer helps the Planner realize the appropriate logic to solve the problem. Since these systems do not have any knowledge of the singularity space of the FR3 or the limitations of the controller, they choose any random point inside the workspace, usually starting from $x=0$, $y=0$ and changing it every iteration based on the feedback from the Simulator. This demonstrates that each feedback mechanism is valuable for achieving an overall robust solution.

For tasks involving a high number of objects, most proposed plans are prone to collisions due to tighter space constraints. The importance of the Simulator's feedback in such cases is highlighted by the \textbf{75}\% improvement it brings to the overall framework in task 4. Task 8 seems to challenge the Planner the most. The combination of arithmetic logic, a higher number of objects in the environment, and workspace constraints means the system needs more feedback loops to resolve such a problem. However, the limit on the number of iterations prevents it from converging on a feasible plan.

\subsubsection{Qualititave Analysis} 
In task planning, it is imperative to observe the actual outputs of a method to understand if it aligns with the input task, environmental conditions, and the robot's constraints. All the modules in our system provide suitable feedback, which can be studied to understand the overall logic and the steps leading to the final proposed plan. For instance, initially, in Figure \ref{fig: test case}, the Planner does not consider that moving an object directly to another object's location would lead to a collision. The Analyzer's feedback corrects this, but the updated plan temporarily places the object in a location unsuitable for the robot. In this case, the Simulator provides feedback to the Planner to further adjust the plan. As seen from this example, our proposed framework increases the quality of the generated plan.

Another example demonstrating the effectiveness of our framework is when the Planner is asked to hand an object to the user, as in tasks 1 and 2. In these cases, the Planner places the object in a random location instead of asking the user for the correct location, despite being prompted to resolve ambiguities by questioning the user. However, with the Analyzer integrated into the framework, such ambiguities are identified in the initial step, prompting the Planner to query the user for clarification. This simple task demonstrates a significant improvement in plan quality by incorporating another agent to review and question the plans generated by the Planner. Moreover, in relatively simple tasks that can be solved solely by the Planner, we observed that the Analyzer typically does not hallucinate, and approves the plan to be passed to the Simulator. Even in instances where the Analyzer hallucinates and sends the plan back to the Planner for revision, the Planner often resubmits the unchanged plan, and the Analyzer approves it upon reevaluation. Hence, introspective dialogue between the two agents also resolves cases where the Analyzer hallucinates.

\begin{table}[!h]
\label{tab:baseline_comp}
\caption{Comparison of the Planning Capabilities of the Baselines and MultiTalk. For the task descriptions please refer to Table \ref{tab:srtable}.}
\centering
\begin{tabular}{@{}cccc@{}}
\toprule
\multirow{2}{*}{\textbf{Task}} & \multicolumn{3}{c}{\textbf{Planning Capabilitiy}}                             \\ \cmidrule(l){2-4} 
                               & \textbf{MultiTalk} & \textbf{ProgPrompt \cite{10161317} } & \textbf{Code-As-Policies \cite{10160591}} \\ \midrule
\textbf{T1}                            & \large{\checkmark}               & \large{\checkmark}                   & \large{\checkmark}                         \\
\textbf{T2}                             & \large{\checkmark}                  & \large{\checkmark}                   & \large{\checkmark}                         \\
\textbf{T3}                            & \large{\checkmark}                  & \xmark                   & \xmark                         \\
\textbf{T4}                              & \large{\checkmark}                  & \xmark                   & \large{\checkmark}                         \\
\textbf{T5}                              & \large{\checkmark}                  & \xmark                   & \large{\checkmark}                         \\
\textbf{T6}                             & \large{\checkmark}                  & \xmark                   & \xmark                         \\
\textbf{T7}                              & \large{\checkmark}                  & \xmark                   & \xmark                         \\
\textbf{T8}                              & \large{\checkmark}                  & \xmark                   & \large{\checkmark}                         \\ \bottomrule
\end{tabular}
\end{table}

\subsection{Baseline Comparison}
We compare MultiTalk with two LLM-based task planning baselines designed for robot manipulation tasks: ProgPrompt \cite{10161317} and Code-As-Policies \cite{10160591}. Both methods are implemented in our environment and tested on the same tasks as the ablation studies in Section \ref{sec: self-abl}. The implementation of ProgPrompt required us to reconstruct the prompts for the system from the information mentioned in their paper, as the original prompts for manipulation tasks were not provided by the authors. Neither baseline aligns plans with respect to the manipulator's capabilities and limitations. Also, they do not consider the precision of the controller or possible object collisions while executing a plan. Hence, we restrict our comparison to logical correctness and alignment with respect to the availability of objects in the environment. We also remove any ambiguity in the task, as these methods lack mechanisms to resolve such issues.

Table \ref{tab:baseline_comp} is generated by running each method on each task five times. Since the objective was to compare the capabilities of different methods, we use a binary measure rather than a measure of accuracy like success rate. If a method generated a feasible plan at least once for a task out of the five executions, it was considered capable of handling the given task. As seen in the table, the interchange tasks (tasks 6 and 7) cause failure in both baselines as the LLM fails to understand the associated logic to properly plan the task. Because of the lack of self-reflecting or critiquing mechanisms in both these approaches, such logical errors go uncorrected. Unlike MultiTalk and Code-As-Policies, ProgPrompt fails in most of the remaining tasks due to its lack of understanding and inability to generate absolute positions for object placement. Overall, it is clearly observed that MultiTalk, with its multiple feedback mechanisms, is able to address logical and grounding aspects more efficiently.

\section{Conclusion}
\label{sec:conclusion}
In this paper, we present a novel methodology for enhancing the robustness and reliability of LLMs for task planning by addressing the associated challenges, such as their propensity to hallucinate, natural language ambiguities in human instructions, uncertainties in the environment, and constraints in the executing agent’s capabilities. Our framework tackles these issues through appropriate introspective and extrospective questioning and dialogue between different modules and the user.

We demonstrate the effectiveness of our framework using robotic arm manipulation tasks and emphasize the necessity of each module through ablation studies. We verify the adaptability of our proposed algorithm to real-world applications by incorporating real robots into our experiments. Additionally, we compare MultiTalk with baseline methods, highlighting its advantages over existing approaches. Our results indicate that our approach mitigates inherent limitations of LLMs in task planning by enhancing the robustness and reliability of their usage in robotic systems. Furthermore, the average cost of the API calls required for a complete task plan generation is only around \$0.15, which makes MultiTalk a scalable solution for robotic task planning.

\section{Limitations and Future Work}

Future work will explore the generalizability of MultiTalk by testing it in more complex environments and with diverse robots. We plan to extend our research to different robotic tasks, particularly for mobile robots, where the Analyzer may have a significant impact due to increased task complexity. Currently, we assume a closed set of objects and available 3D models for the Simulator. Future efforts will address this limitation by developing methods to generate approximate object models on-the-fly for open-set simulations. Although the planning loop can be run multiple times during plan execution to account for dynamic environments, future work will focus on adapting plans for rapidly changing environments.








\bibliographystyle{IEEEtran}
\bibliography{references}

\end{document}